%% file: acl_latex.tex
\documentclass[11pt]{article}
\usepackage[preprint]{acl}
\usepackage{times}
\usepackage{latexsym}
\usepackage[T1]{fontenc}
\usepackage[utf8]{inputenc}
\usepackage{microtype}
\usepackage{float}
\usepackage{inconsolata}
\usepackage{amsmath}
\usepackage{graphicx}
\usepackage{amsmath,amssymb,amsfonts,amsthm,latexsym}
\usepackage{booktabs}
\usepackage{array}
\newcolumntype{L}[1]{>{\raggedright\arraybackslash}p{#1}}

\DeclareUnicodeCharacter{1ECA}{\d{I}}
\DeclareUnicodeCharacter{1ECB}{\d{i}}
\DeclareUnicodeCharacter{1ECC}{\d{O}}
\DeclareUnicodeCharacter{1ECD}{\d{o}}
\DeclareUnicodeCharacter{1EE4}{\d{U}}
\DeclareUnicodeCharacter{1EE5}{\d{u}}

\newcommand{\promptbox}[2]{%
  \par\medskip\noindent
  \fbox{\begin{minipage}{\dimexpr\columnwidth-2\fboxsep-2\fboxrule\relax}
    \raggedright\small\textbf{#1}\par\smallskip
    \footnotesize\ttfamily #2
  \end{minipage}}\par\medskip}

\usepackage{xcolor}
\definecolor{igbogreen}{HTML}{0F6B45}
\definecolor{englishgray}{HTML}{3F444B}
\definecolor{loopcolor}{HTML}{E2622C}
\definecolor{loopink}{HTML}{B8441A}
\definecolor{bartrack}{HTML}{E6E5DF}
\definecolor{statgray}{HTML}{6B6A66}
\newcommand{\loopword}[1]{\textcolor{loopink}{#1}}
\newlength{\loopbarwidth}
\newcommand{\loopbar}[1]{\raisebox{0.15ex}{\textcolor{loopcolor}{\rule{#1\loopbarwidth}{5pt}}\textcolor{bartrack}{\rule{\dimexpr\loopbarwidth-#1\loopbarwidth\relax}{5pt}}}}

\AtBeginDocument{\nolinenumbers}

\title{{\LARGE Do Language Models Understand Igbo Culture?}\\[4pt]{\large\itshape Beyond Multiple Choice}}

\author{%
  \begin{minipage}{\textwidth}
    \centering
    \large
    Blessing Ngozi Uzor$^{1}$\hspace{1.4em}Valdini Douglace Lemofouet$^{1}$\hspace{1.4em}Paula Chikaodinaka Anyanwu$^{1}$ \\[2pt]
    \makebox[\linewidth][c]{Mmesoma Peace Orji$^{2}$\hspace{1.4em}Danielle Blanche Kapsa$^{1}$\hspace{1.4em}Tassallah Abdullahi$^{3}$\hspace{1.4em}Clemencia Siro$^{4}$} \\[2pt]
    Idris Abdulmumin$^{5}$\hspace{1.4em}Seid Muhie Yimam$^{6}$\hspace{1.4em}Shamsuddeen Hassan Muhammad$^{7,8}$ \\[6pt]
    \normalsize\mdseries
    \mbox{$^{1}$African Institute for Mathematical Sciences (AIMS), Cameroon}\quad
    \mbox{$^{2}$University of Nigeria, Nsukka}\quad
    \mbox{$^{3}$Brown University}\quad
    \mbox{$^{4}$Centrum Wiskunde \& Informatica}\quad
    \mbox{$^{5}$University of Pretoria}\quad
    \mbox{$^{6}$University of Hamburg}\quad
    \mbox{$^{7}$Bayero University Kano}\quad
    \mbox{$^{8}$Imperial College London}
  \end{minipage}%
}

\usepackage{hyperref}
\hypersetup{
    pdfinfo={
        Title={Do Language Models Understand Igbo Culture? Beyond Multiple Choice},
        Author={Blessing Ngozi Uzor, Valdini Douglace Lemofouet, Paula Chikaodinaka Anyanwu, Mmesoma Peace Orji, Danielle Blanche Kapsa, Tassallah Abdullahi, Clemencia Siro, Idris Abdulmumin, Seid Muhie Yimam, Shamsuddeen Hassan Muhammad},
        Subject={},
        Keywords={}
    }
}

\begin{document}
\maketitle
\begin{abstract}
The value of language models for any community depends on whether a model actually understands the culture of the people it intends to serve. We use the Igbo language as a case study. Existing benchmarks test translated Western ideas, lacking depth. To address this, we introduce CAB-Igbo, a benchmark developed from  indigenous cultural sources and verified by native speakers across six domains. We evaluated diverse models under varying conditions, incorporating regional personas. Performance was assessed using multiple-choice accuracy and human evaluations across Epistemic Fidelity, Representational Richness, Linguistic Quality, and Pragmatic Proficiency. Results reveal a stark dissociation between language fluency and cultural knowledge. Through statistical testing, we show that for most models, the severe lack of Epistemic Fidelity and Representational Richness is already present in English and does not significantly worsen when switching to Igbo. Instead, native language prompting mainly degrades Linguistic Quality and Pragmatic Proficiency. Gemini 2.0 Flash is the only model whose cultural knowledge scores improve under Igbo prompting. We find that current models struggle to demonstrate the knowledge needed for true alignment with Igbo, and multiple-choice formats allow pattern-matching to appear as understanding. Furthermore, automated judges agree with humans on linguistic quality but fail on epistemic fidelity. Ultimately, cultural evaluation must be built from community knowledge, establishing human evaluation as the gold standard to prevent misrepresenting low-resource communities globally.
\end{abstract}

\section{Introduction}
Language goes beyond vocabulary; it functions as the historical archive of a people's shared experiences. As a community navigates similar circumstances generation after generation, a distinct culture evolves, holding the community's ethical, moral, and aesthetic standards. When a society is culturally healthy, there is a natural seamless alignment uniting the words people speak, the symbols they write, and the reality of their daily interactions. But what happens when this harmony is disrupted? Historically, this fracture began within the colonial education system, which separated the language of conceptual learning from the language of everyday life. Today, we face a new frontier of this same alienation: the Large Language Model.

Language models are increasingly deployed globally to preserve heritage and process information \citep{Zhao2025EvaluationOL}, yet they are overwhelmingly trained on English and other high-resource languages \citep{resnik2025large}. We are told that these frontier models are "multilingual," that they can speak to us in our local languages. But linguistic fluency must not be confused with cultural fidelity. As an act of resistance against this modern alienation, this research investigates whether Large Language Models genuinely grasp the depths of African cultures, centering on the Igbo language and its people. Spoken by roughly 30 million individuals in Southern Nigeria, Igbo carries a profound cultural heritage anchored in vibrant oral traditions and complex kinship networks. When an artificial intelligence processes fluent Igbo but maps it onto Western norms or pan-African stereotypes, it performs significant representational harm. It preserves surface meaning but strips away the deep cultural ontology embedded in the expression.

Cultural misalignment goes beyond bad test scores, threatening real-world sectors like education, public services, and safety \citep{Gallegos2024}. For communities like the Igbo, whose knowledge lives in oral traditions, algorithmic erasure replaces a vibrant history with generic foreign defaults \citep{joshi-etal-2020-state, Nekoto2020, Naous2024, tonga-etal-2026-llms}. Cultural alignment is not an optional add-on. It is an absolute requirement for building safe and equitable AI \citep{dev2026unified, zhou2025culture, ouyang2022training, Bai2022ConstitutionalAH, ryan2024unintendedimpactsllmalignment, Orlowski2025TooMA}.

To measure this gap between speaking a language and knowing a culture, this study introduces the Cultural Alignment Benchmark for Igbo (CAB-Igbo). Built from the ground up using verified knowledge from the Igbo community across six domains, this benchmark rejects the imposition of outside templates. By evaluating eight state-of-the-art models under varying linguistic and persona conditions, and measuring responses against four dimensions (Epistemic Fidelity, Representational Richness, Linguistic Quality, and Pragmatic Proficiency), we seek to uncover whether these models truly reflect Igbo values or merely mimic our words while remaining strangers. We treat linguistic quality as a distinct dimension from epistemic fidelity to disentangle surface-level language proficiency from culturally grounded knowledge representation.

\subsection{Research Questions}
This raises important questions:
\begin{enumerate}
    \item \textit{How well do language models understand Igbo culture, and what misalignment patterns appear in current models?}
    \item \textit{Does the evaluation language (English or Igbo) affect model performance, and does fluency imply cultural competence?}
    \item  \textit{Does model scale determine cultural alignment, or is training data specificity the most important?}
    \item \textit{Do models trained on African data show better cultural alignment than English-centric ones?}
\end{enumerate}

\subsection*{Our Contributions}
\begin{enumerate}
    \item \textit{We introduce CAB-Igbo, a novel, bilingual benchmark designed to measure cultural alignment in the Igbo language. Moving beyond surface-level translation, it evaluates deep cultural practices across six semantic domains of Igbo life.}
    \item \textit{We empirically demonstrate across eight large language models that high multiple-choice accuracy relies heavily on pattern matching, masking a severe inability to generate culturally grounded reasoning in open-ended settings.}
    \item \textit{We show that while LLM-as-a-judge correlates strongly with human evaluators on linguistic fluency, it systematically fails to assess deep cultural proxies.}
\end{enumerate}
\section{Related Works}
Early NLP evaluation relied on surface metrics. As models moved to general-purpose foundational models, evaluation has shifted towards full scope benchmarks. Tools like Global MMLU \citep{Singh2024} and BIG-Bench measure general capabilities across domains. Model scale improved fluency but did not produce cultural depth. To fill gaps in multilingual evaluation, researchers used translation. Tools like Global MMLU \citep{Singh2024} translated English benchmarks into other languages, including African ones. But language transfer is not culture transfer. Translating Western tests into African languages does not check for African understanding; it evaluates whether a model can process Western cultural logic in an African linguistic form. This flattens differences and shows that speaking many languages is not the same as knowing local ways. Researchers then turned to surveys \citep{Adilazuarda2024}. Benchmarks like CDEval \citep{Wang2023CDEvalAB} and Hofstede-based analyses \citep{Masoud2023CulturalAI} use the World Values Survey and Hofstedes dimensions. They ask models multiple-choice questions from one groups point of view. Reducing dynamic cultural practices to fixed categorical value dimensions fails to capture situated social realities. Culture is not a checklist. Furthermore, recent empirical studies show that persona prompting across different languages is highly unstable \citep{Khan2025RandomnessNR, AlKhamissi2024}. Models fall back on shallow stereotypes rather than reasoning about culture. Translation and surveys force outside templates. Native benchmarks respond by using datasets like BLEnD \citep{Myung2024BLEnDAB} and CulturalBench \citep{chiu-etal-2025-culturalbench}, which let locals ask their own questions about their culture and everyday life. But BLEnD has items not true to local Northern Nigerian daily living, e.g., ``What is the main dish for Thanksgiving in Northern Nigeria?'' and ``What do people do to celebrate Halloween in Northern Nigeria?'' Neither practice exists in these communities. Adaptability benchmarks such as NormAd \citep{Rao2025} use real-life scenarios. Recent benchmarks prove that cultural evaluation requires native curation. For example, \citet{myculture2024} evaluates Malaysian culture using paired closed and open-ended formats, finding a massive gap between multiple-choice guessing and generative reasoning. We build on MyCulture's approach, expanding it with our diagnostic distractor taxonomy and intra-cultural personas. Similarly, XCR-Bench introduces a multi-task framework to evaluate cross-cultural adaptation, demonstrating that models struggle with implicit social norms even when surface-level translations are accurate \citep{xcrbench2024}. Recent large-scale studies on persona prompting have also revealed that injecting demographic variables often yields unstable or negative effects rather than improving cultural reasoning \citep{Khan2025RandomnessNR, AlKhamissi2024}. Finally, datasets like NRITYAM highlight the necessity of expert, community-validated datasets for evaluating culturally grounded knowledge in specific domains \citep{nrityam2024}.

\section{CAB-Igbo Dataset}
CAB‑Igbo is a bilingual (English‑Igbo) evaluation task. It consists of 296 unique cultural question paradigms. Because each paradigm is translated and evaluated in both English and Igbo, this forms a core dataset of 592 items. Every question is asked in two paired formats. As a multiple-choice question (MCQ), it has a cultural premise and four answer choices, which include diagnostic distractors such as Western defaults, ethnolinguistic substitutions, and Pan‑African generalisations. As an open-ended short-answer question (SAQ), the same question is asked without any answer choices, and the model must explain the cultural reasoning in its own words in two to four sentences. All instances are drawn from authentic Igbo community sources. Native speakers verify every question. The correct answer depends on culturally grounded context, not surface‑level pattern matching or simple language fluency. Because the SAQ removes the answer choices, CAB-Igbo tests not only whether a model can recognise the right answer but also whether it can explain it. This makes CAB‑Igbo a valuable benchmark for evaluating whether an LLM can achieve true cultural alignment and incorporate authentic indigenous knowledge in low‑resource language tasks.

\subsection{Dataset Construction}
CAB-Igbo is built from scratch using expert-guided primary sources rather than web scraping, which reduces the likelihood of direct pre-training data leakage. However, we do not claim this eliminates leakage risk entirely: common proverbs or widely known cultural practices referenced in our items may still overlap with content present in pre-training corpora. Our diagnostic distractors and open-ended reasoning prompts are novel constructions not found in any public source, which limits the extent to which simple memorization alone could explain strong performance. The dataset captures regional nuances from Imo and Ebonyi states. To guarantee quality, we used a participatory method, checking every item for factual accuracy, language flow, and cultural fit. If the community did not reach consensus, the item was removed. 

\begin{figure*}[htbp]
    \centering
    \input{fig_samples.tex}
    \caption{Samples from the CAB-Igbo Dataset. Questions are grounded in authentic cultural domains and evaluate deep epistemic fidelity rather than mere linguistic translation.}
    \label{fig:cab_samples}
\end{figure*}

\subsection{Evaluator Requirements}
We engaged expert Igbo linguists. We also leveraged the expertise of native Igbo-speaking evaluators with community-level cultural knowledge. They provided authentic cultural expertise. The linguist rigorously verified every line of Igbo in the benchmark. Ensuring it met the highest language standard. Native speakers validated the dataset using a participatory method. This ensured the benchmark was rooted in real community knowledge rather than outside guesses or machine‑generated content. Each item was reviewed for factual accuracy, language quality, and cultural fit. Any items lacking consensus among the speakers were modified or removed.

\subsection{Regional and Domain Selection}
To ensure broad and authentic coverage of the Igbo worldview \citep{Nwoye2011}, we structured the benchmark around six key domains. These are: Family and Kinship, Festivals and Religion, Food and Cuisine, Funerals and Burial, Proverbs and Oral Tradition, and Social Customs and Etiquette. We created cultural personas based on two distinct Igbo states: Imo and Ebonyi. Imo State was prioritized because it serves as the language reference point for Igbo Izugbe, or Standard Central Igbo. It is also highly represented in digital resources. Ebonyi State was selected for a different reason. It features an underrepresented dialect. This allowed us to introduce language variation and test whether models could actually differentiate intra‑Igbo regional identities, rather than just matching standard digital patterns.

\subsection{Question Formatting and Data Statistics}
We considered writing separate items for different testing formats. However, we chose a stricter approach. Following a similar evaluation methodology utilized in recent cultural benchmarks \citep{myculture2024}, we ask the exact same question in two distinct formats.  CAB‑Igbo presents 296 core cultural paradigms as multiple‑choice questions (MCQs) and as open‑ended short answer questions (SAQs) (see Figure \ref{fig:cab_samples} for examples). This paired design tests whether models can do more than surface‑level pattern matching. It reveals whether a model can correctly pick the right option and still explain the cultural reasoning behind it without help from the provided choices. This process produced 296 unique question pairs per language. Across both English and Igbo, this creates a foundational dataset of 592 unique items. Because each is evaluated in both multiple-choice and open-ended formats, it yields 1,184 distinct testing conditions before any persona constraints are applied. We did not use broad, country‑level distributions. Instead, CAB‑Igbo captures intra‑cultural variations. The 296 core paradigms scale into a comprehensive evaluation grid. We multiply them across three linguistic prompt types: English, English‑Igbo, and Igbo‑Igbo. We also include 11 persona conditions and two answer permutations. The exact text of every prompt template used in our experiments is given in Appendix~\ref{app:prompt_templates}. This architecture generates tens of thousands of test instances. During evaluation, it yielded 78,144 closed‑form responses under the original answer ordering, and 1,416 open‑ended generations on a 20\% sample of the benchmark (59 questions), of which 1,415 were human‑scored.

\section{Experimental Setup}
\label{sec:experimental_setup}
We conducted experiments across 8 large language models. We grouped them into four distinct categories to assess the impact of scale, training data, and multilingual focus. The first group is an Africa‑specific open model, AfriqueQwen 14B \citep{Yu2026}. The second group includes multilingual open models: LLaMA 3.3 70B and LLaMA 3.1 8B. The third group includes general‑purpose open models: LLaMA 3 8B Lite, GPT‑OSS 120B, and GPT‑OSS 20B. The fourth group includes closed‑weight frontier models: GPT‑4o mini and Gemini 2.0 Flash. All experiments used zero temperature and greedy decoding. This enforced deterministic and factual outputs. For GPT-OSS 120B and GPT-OSS 20B, responses were limited to 300 output tokens for MCQs and 500 for SAQs. These are reasoning models whose internal reasoning counts toward this limit, and in our first run the reasoning often used up the whole budget, so the model returned an empty MCQ response: 5,640 of 19,536 responses for GPT-OSS 120B and 11,826 of 19,536 for GPT-OSS 20B. We therefore ran a second pass for GPT-OSS 120B in which only these empty items were queried again with a limit of 4,000 tokens, using the same provider, prompts, decoding settings, and option orderings. Every re-queried item returned a valid option letter, and no response from the first run was changed. GPT-OSS 20B could not be re-queried in the same way because the provider no longer offered the model on its shared service; its MCQ results therefore come from the first run, and its empty responses are scored as incorrect. We conducted experiments across three linguistic prompt types. Type 1 is English‑English. Type 2 is English‑Igbo. Type 3 is Igbo‑Igbo. We evaluated the models using two paired strategies. The first is multiple‑choice questions (MCQs). The second is open‑ended short answer questions (SAQs). For MCQs, we assigned alphabetical labels to the answer choices: A, B, C, and D. We ran each MCQ through two answer permutations to test for positional bias. The first permutation kept the original written order. The second used a fixed-seed shuffled order. While full permutation sweeps ($4! = 24$ combinations) would provide a more exhaustive estimate of positional bias, computational and API constraints limited us to these two conditions, which were sufficient to establish the baseline presence of bias. For the open-ended approach, we prompted models to generate a free-form explanation. We restricted responses to two to four sentences. A native human judge then evaluated these responses across four constructs. These are Epistemic Fidelity, Representational Richness, Linguistic Quality, and Pragmatic Proficiency. This allowed us to assess thick cultural reasoning \citep{Geertz1973}. Cultural knowledge varies across locations and demographics. We hypothesized that providing specific socio‑geographical context can affect a model's reasoning ability. To test this, we complemented our zero‑shot baseline experiments with ten different persona conditions. These represented five distinct socio‑economic roles. Examples include a young teenage boy, a married woman, and a professor. We mapped these roles across two specific Igbo states: Imo and Ebonyi. This setup allowed us to measure whether models could differentiate intra‑Igbo regional identities. It also tested whether they would default to homogeneous generalizations.

\subsection{Statistical Methodology}
To rigorously evaluate the significance of performance differences across conditions, we applied two distinct statistical approaches tailored to the evaluation formats. For multiple-choice questions (MCQs), which yield paired binary outcomes (correct vs. incorrect on the same item under two conditions), we employed McNemar's test with Yates' continuity correction. This was used to assess the significance of prompt type degradation (English vs. Igbo), positional bias (original vs. shuffled option ordering), and persona effects (baseline vs. demographic persona). For the open-ended short answer questions (SAQs), which were scored by human evaluators on ordinal 0-2 scales across four constructs, we computed 95\% bootstrap confidence intervals (using 10,000 resamples) on the mean within-item score differences ($\Delta$) between the English baseline and the Igbo prompt conditions. Differences were considered statistically significant if the 95\% confidence interval did not include zero. This non-parametric approach avoids strict distributional assumptions and provides robust estimates for the shift in generative cultural reasoning.

\section{Analysis}
\subsection*{Result by Categories}
We evaluate the performance of eight selected language models on CAB‑Igbo. We provide a detailed analysis across six cultural domains and three linguistic prompt types. We further categorize each sample by its regional condition. This allows us to assess intra‑Igbo cultural differentiation between Imo and Ebonyi. We structure our analysis in three parts. First, we examine baseline MCQ accuracy. This establishes how well models perform at surface‑level cultural recognition. Second, we evaluate performance on open‑ended short answer questions. We score these using our multi‑dimensional human evaluation rubric and an LLM‑as‑a‑judge protocol. This assesses their capacity for authentic cultural reasoning. Third, we synthesize these findings. This synthesis exposes the central finding of our study: the gap between multiple‑choice pattern matching and true cultural generation.

\subsection{MCQs}
\subsection*{Linguistic Prompt Type Performance}
\label{sec:prompt_type}
\begin{figure*}[t]
\centering
\includegraphics[width=\textwidth]{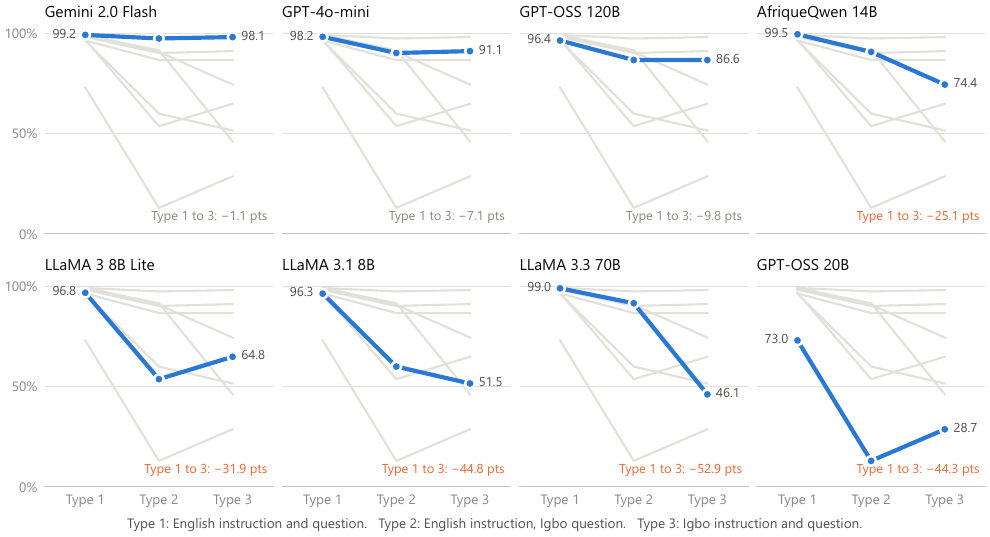}
\caption{Prompt Degradation: The collapse of cultural reasoning across English and Igbo prompt types.}
\label{fig:prompt_degradation}
\end{figure*}

We tested three conditions: Type 1, Type 2, and Type 3. For Gemini 2.0 Flash, the effect is small, dropping by only 1.1 percentage points. GPT-4o mini drops more with 7.1 points. Interestingly, it does slightly better on full Igbo than when it receives mixed English–Igbo prompts. LLaMA 3.3 70B scores 99.0\% on Type 1 but then falls to 46.1\% on Type 3. This is a massive drop of 52.9 points. This is the "fluent but foreign" effect described by \citet{Agarwal2025FluentBF}, where models seem fluent but misaligned (see Figure \ref{fig:prompt_degradation}). AfriqueQwen 14B is as sensitive to language as LLaMA 3.3 70B, but steadier, dropping about 25.1 points from 99.5\% on Type 1 to 74.4\% on Type 3. McNemar's tests on the zero-shot baseline items (Type~1 vs Type~2, original ordering; Table~\ref{tab:mcnemar_results}) confirm that the English-to-Igbo degradation is statistically significant for every model tested (all $p<0.05$, $\chi^2$ ranging from 5.14 for Gemini 2.0 Flash to 164.28 for GPT-OSS 20B), establishing that the ``fluent but foreign'' collapse is not an artefact of a few unlucky items but a systematic effect across the benchmark.

\subsection*{Effect of Persona}
\label{sec:persona}
Table~\ref{tab:persona_baseline} asks a more direct question: does persona prompting help or hurt the model? It compares zero-shot baseline accuracy with mean accuracy across ten persona conditions for each model. GPT-4o mini gains 1.6 points under persona conditions, a modest but consistent improvement. For AfriqueQwen 14B and LLaMA 3.3 70B, persona actively lowers accuracy by introducing demographic noise into the given prompt. GPT-OSS 120B is almost unaffected, changing by only 0.3 points. GPT-OSS 20B loses 8.5 points, the biggest decline we saw, though its many empty responses (Section~\ref{sec:experimental_setup}) blur the line between persona effects and output breakdown. Where other models either stay flat or decline, LLaMA 3.1 8B gains 10.4 points under persona conditions, the only large positive shift we saw in Table~\ref{tab:persona_baseline}. Table~\ref{tab:persona_detail} compares how each model responds to different demographic personas. Persona identity has little power over cultural accuracy. LLaMA 3.1 8B shows the widest gap at 4.7 points, between the Professor and Widowed Woman personas. Frontier models differ by at most 0.8 points. Our persona experiments also align with recent large-scale studies showing that demographic prompting often hurts rather than helps \citep{Khan2025RandomnessNR, AlKhamissi2024}. A persona prompt cannot magically inject missing cultural knowledge. As observed in the data regarding the inter-persona delta, LLaMA 3.1 8B improves under every persona condition, gaining between 8.1 to 12.8 percentage points above its zero-shot baseline. The Professor persona provides the biggest gain at 12.8 points. However, this does not prove that the model has actual lived cultural knowledge; instead, the persona works as a simple prompt that pushes the model to answer more carefully. AfriqueQwen 14B shows the exact reverse. Every persona reduces its accuracy by roughly five to six points from its 93.2\% baseline. Its Africa-specific training may have optimised it for direct questions without additional persona framing. McNemar's tests confirm this asymmetry statistically (Table~\ref{tab:mcnemar_results}). Each persona condition was compared with the zero-shot baseline on the same 888 question and prompt type pairs in the original ordering, the same items that underlie Table~\ref{tab:persona_baseline}. Persona effects are significant for all ten personas for AfriqueQwen 14B (drops; $\chi^2$ from 21.33 to 36.01), LLaMA 3.1 8B (gains; $\chi^2$ from 20.05 to 69.14), and GPT-OSS 20B (drops; $\chi^2$ from 17.33 to 46.52), for nine of ten personas for LLaMA 3.3 70B (drops), and for seven of ten for GPT-4o mini (gains). For Gemini 2.0 Flash, GPT-OSS 120B, and LLaMA 3 8B Lite, no persona condition differs significantly from baseline (all $p>0.05$). Persona framing therefore shifts accuracy systematically for most models, but the direction depends on the model, and no persona produces a consistent benefit across models.

\subsection*{State-Level Analysis}
\subsection*{Performance Gap Across States}
\label{sec:state_level}
Table~\ref{tab:state} records accuracy for each model under two state-level persona conditions: Imo and Ebonyi. Can models tell Imo from Ebonyi at a cultural level? If models held regional knowledge, the Imo condition should significantly outperform Ebonyi because it represents the standard \textit{Igbo Izugbe} heavily found in digital texts. Instead, as the data shows, the gaps between Imo and Ebonyi range from a mere 0.1 to 0.6 points, indicating that performance lines run nearly parallel. Models treat state labels as empty markers: they register the words but encode no difference in dialect, culture, or local knowledge. MCQs may be the wrong tool to detect persona effects and state-level differences. Because the correct option is the same for both Imo and Ebonyi, a model needs only general plausibility to score well, requiring no true regional knowledge.

\subsection*{The Effect of Positional Bias}
\label{sec:position_bias}
Descriptively, six of the eight models showed a raw point drop under shuffled ordering. McNemar's tests (with continuity correction) on the zero-shot baseline items, pairing the original and shuffled ordering of each of the 888 question and prompt type items, show that the drop is significant for four models: GPT-4o mini ($\chi^2=16.60$, $p<0.001$), LLaMA 3.1 8B ($\chi^2=11.52$, $p<0.001$), Gemini 2.0 Flash ($\chi^2=5.26$, $p=0.022$), and LLaMA 3.3 70B ($\chi^2=4.34$, $p=0.037$). LLaMA 3 8B Lite is borderline ($\chi^2=3.36$, $p=0.067$). For AfriqueQwen 14B and both GPT-OSS models, the difference is not significant (Table~\ref{tab:mcnemar_results}). Positional bias is therefore present in several models, including both frontier models, which suggests that part of their MCQ accuracy depends on option order rather than on cultural knowledge.

\begin{figure*}[t]
    \centering
    \includegraphics[width=\textwidth]{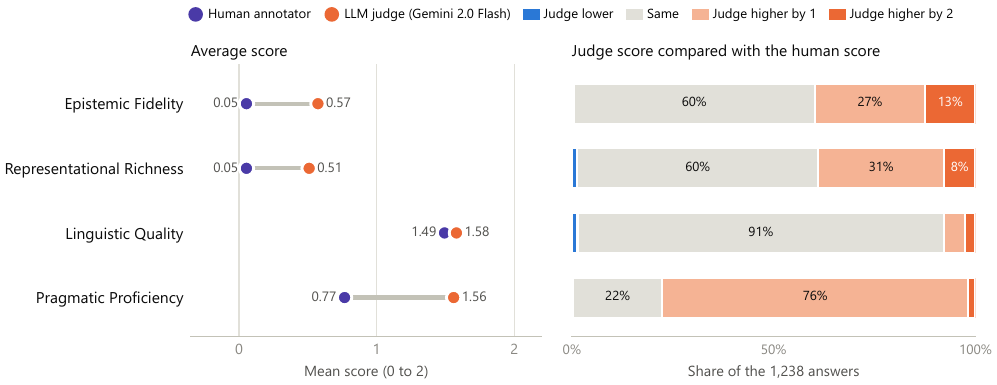}
    \caption{The Limits of Automated Evaluation: Comparing human evaluator scores against the Gemini 2.0 automated judge. The automated judge significantly overestimates Epistemic Fidelity and Representational Richness compared to native human experts.}
    \label{fig:human_vs_llm}
\end{figure*}

\subsection{Human Evaluation}
\label{sec:open_ended}
\subsection*{Open-Ended Evaluation Overview}
We graded open-ended answers on four criteria. Each measure was scored between 0 to 2. The four are Epistemic Fidelity (EF), Representational Richness (RR), Linguistic Quality (LQ), and Pragmatic Proficiency (PP). A response has a maximum of 8 points total. We evaluated a 20\% sample of the benchmark (59 questions), giving 1,416 generations, of which 1,415 were scored. We tested all eight models across all three prompt types (under zero-shot conditions only). Each construct measures something different: EF and RR test how deep and true the cultural knowledge is, LQ tests whether the Igbo grammar flows and is readable, and PP tests whether the answer fits the required situation.

\subsection*{Human Evaluation Scores by Model}
\label{sec:human_eval_model}
\begin{figure*}[t]
    \centering
    \includegraphics[width=\textwidth]{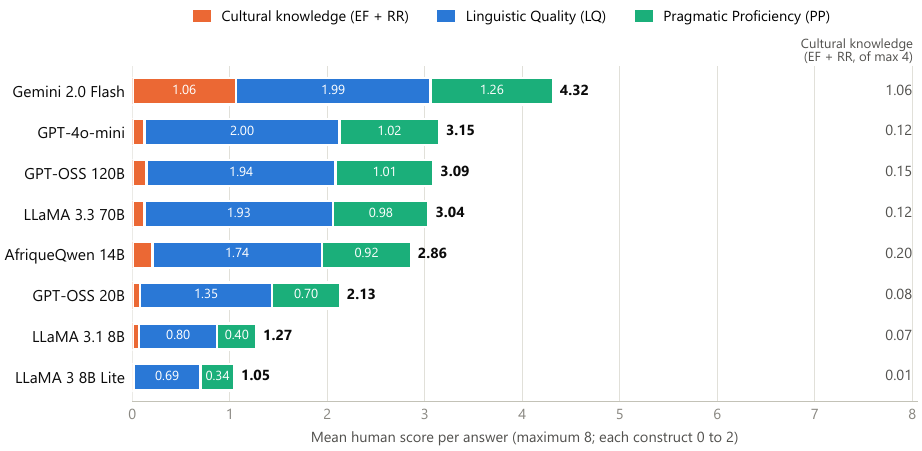}
    \caption{Aggregate Human Evaluation Scores. Gemini 2.0 Flash significantly outperforms all other models, while smaller open-weight models suffer near-complete breakdown in culturally grounded contexts.}
    \label{fig:model_leaderboard}
\end{figure*}

Table~\ref{tab:human_overall} reports mean scores per model aggregated across all three prompt types (the aggregate scores are also summarized visually in Figure \ref{fig:model_leaderboard}). Gemini 2.0 Flash stands out from all other models. Its mean total score is 4.316 out of 8. Its EF and RR scores are 0.531. Gemini's knowledge scores are five times higher than the next best model. Four models sit in the middle tier: GPT-4o mini, GPT-OSS 120B, LLaMA 3.3 70B, and AfriqueQwen 14B \citep{Yu2026}. Their total scores range from 2.859 to 3.147, making them nearly identical. Their LQ scores reach near maximum (1.740 to 2.000), meaning their Igbo sounds smooth and correct. But EF and RR are stuck exceptionally low, between 0.062 to 0.102. In this middle group, size and training source do not matter for cultural knowledge. The lowest tier contains three models: GPT-OSS 20B, LLaMA 3.1 8B, and LLaMA 3 8B Lite. Their total scores range from 1.045 to 2.131. LLaMA 3 8B Lite scores just 1.045 on average across all conditions. We note that EF and RR scores are identical not only on average but for every individual response in our sample. We argue this reflects a genuine property of cultural knowledge rather than rubric redundancy: representational richness without epistemic accuracy is not depth but elaboration on a false premise. A response cannot meaningfully demonstrate rich, nuanced cultural understanding (RR) while being factually or culturally incorrect (low EF), since the richness itself would be built on a mistaken premise. In our benchmark, where each item targets a specific, verifiable cultural referent, this dependency manifests as near-total convergence between the two constructs. This suggests EF may function as a gating condition for RR in culturally grounded evaluation, a relationship worth testing explicitly in future work with items designed to decouple the two (e.g., responses that are accurate but sparse, versus elaborate but wrong).

\subsection*{Prompt Type Effects in Open-Ended Evaluation}
\label{sec:prompt_type_human}
Table~\ref{tab:type_degradation} shows mean scores by model and prompt type. Table~\ref{tab:bootstrap_results} reports bootstrap 95\% confidence intervals (10,000 resamples) on the within-item difference between Type~1 and Type~2 scores for each construct. For six of the eight models (AfriqueQwen 14B, GPT-4o mini, GPT-OSS 120B, GPT-OSS 20B, LLaMA 3.3 70B, and LLaMA 3 8B Lite), the change in Epistemic Fidelity and Representational Richness from English to Igbo prompting is \textbf{not} statistically significant (95\% CI includes zero). In contrast, Linguistic Quality drops significantly for five models (AfriqueQwen 14B, GPT-OSS 20B, LLaMA 3.3 70B, LLaMA 3.1 8B, and LLaMA 3 8B Lite) and Pragmatic Proficiency for three (GPT-OSS 20B, LLaMA 3.1 8B, and LLaMA 3 8B Lite). GPT-4o mini and GPT-OSS 120B show no significant change on any construct. This dissociation is informative: for most models, low EF and RR are not a consequence of switching languages, but reflect a knowledge gap already present when models are prompted in English. Igbo prompting introduces a separate failure, a decline in fluent generation, rather than causing the cultural knowledge gap itself. The non-significance of EF and RR changes should not be read as a positive result. It indicates that the absence of cultural knowledge is already present in English and is not a translation artefact. For LLaMA 3 8B Lite, it also reflects a floor effect, since its Type~1 EF is already 0.017. Two models depart from this pattern. LLaMA 3.1 8B is the only model whose EF and RR drop significantly ($\Delta=0.10$), alongside much larger LQ and PP collapses, consistent with its breakdown under Igbo input (see Section~\ref{sec:error_analysis}). Gemini 2.0 Flash moves in the opposite direction: its EF and RR rise significantly under Igbo prompting, from 0.237 on Type~1 to 0.576 on Type~2 and 0.780 on Type~3 ($\Delta=-0.34$, 95\% CI $[-0.58, -0.10]$ for Type~1 vs Type~2), and its PP rises as well ($\Delta=-0.19$). Gemini 2.0 Flash is the only model whose cultural knowledge scores improve when the question is posed in Igbo.

\subsection{LLM-as-a-Judge Evaluation}
\label{sec:llm_judge}
Table~\ref{tab:judge_scores} reports mean scores assigned by Gemini 2.0 Flash as an automated judge to the open-ended responses of seven models. Each model was judged on 177 responses (59 questions across all three prompt types), giving 1,239 evaluations, of which 1,238 have a matching human score. We selected Gemini 2.0 Flash as the judge due to its strong multilingual instruction-following capabilities. Crucially, to ensure strict independence and completely avoid self-preference bias or style alignment distortion, we deliberately excluded Gemini 2.0 Flash's own generations from this adjudication phase. Thus, for the LLM-as-a-judge experiment, Gemini functioned entirely as a distinct evaluator that was not among the tested systems being scored. Column $n$ records the number of matched responses per model. GPT-OSS 120B earns the highest judge total and the highest EF, while GPT-4o mini achieves a perfect 2.000 on LQ and PP across all responses. The two smallest models, LLaMA 3.1 8B and LLaMA 3 8B Lite, score lowest on every construct. Crucially, when judge scores are placed alongside the human scores for the same 1,238 responses (Table~\ref{tab:ira}), the automated judge awards higher scores than the human annotator on every construct (visualized in Figure \ref{fig:human_vs_llm}). The gap is smallest for Linguistic Quality. The largest discrepancies appear in Epistemic Fidelity, Representational Richness, and Pragmatic Proficiency.

\subsection*{Inter-Rater Agreement}
\label{sec:ira}
Table~\ref{tab:ira} reports inter-rater agreement between the human annotator and the Gemini 2.0 Flash judge across 1,238 matched response pairs, using Cohen's weighted $\kappa$ and Spearman's $\rho$. Linguistic Quality shows near-perfect agreement ($\kappa_w = 0.833$, $\rho = 0.885$, exact agreement 90.9\%), confirming that surface fluency is reliably measurable by automated judges. Epistemic Fidelity and Representational Richness show poor agreement ($\kappa_w < 0.115$). The judge is systematically too lenient on deep cultural content. Pragmatic Proficiency presents a split result: $\rho = 0.883$ suggests strong rank-ordering, but $\kappa_w = 0.261$ and just 22.1\% exact agreement reveal that the judge scores PP 0.792 points higher on average. These results establish human evaluation as the gold standard for cultural depth: automated judges can reliably assess fluency but fail on epistemic fidelity.

\begin{figure*}[t]
    \centering
    \includegraphics[width=\textwidth]{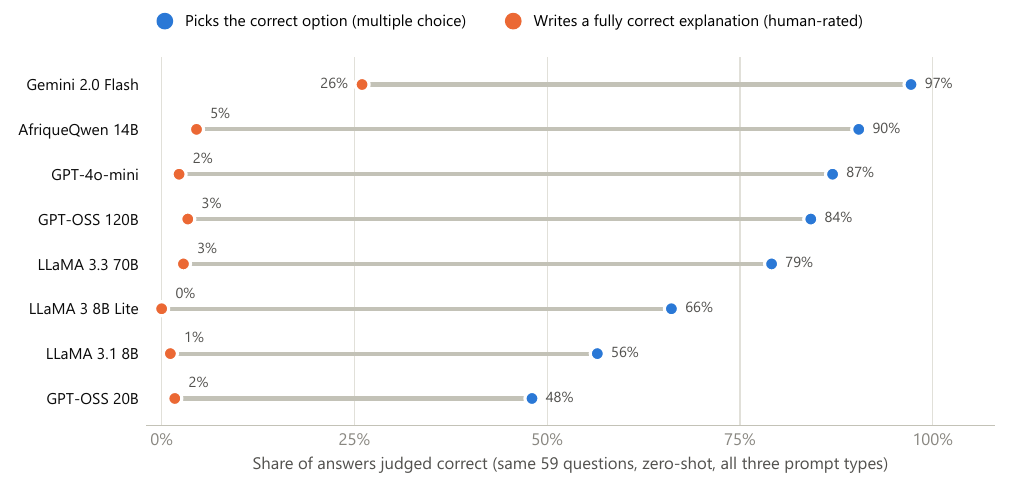}
    \caption{The gap between recognising and explaining Igbo culture. For each model, on the same 59 questions in the zero-shot condition across all three prompt types (177 items per model; 176 open-ended answers for GPT-OSS 20B): the share of MCQ items answered correctly (original ordering) and the share of open-ended answers rated fully correct by the human annotator (Epistemic Fidelity $=2$).}
    \label{fig:gap}
\end{figure*}

\begin{figure}[t]
    \centering
    \includegraphics[width=\columnwidth]{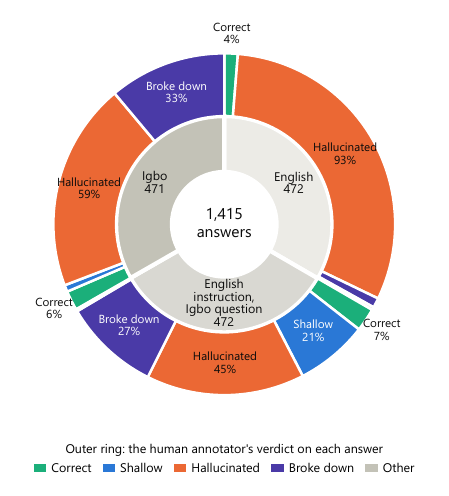}
    \caption{Primary human verdict on each of the 1,415 scored open-ended answers, by prompt type. The verdict is the first error label the annotator assigned; Correct covers answers labelled Culturally Accurate or with no error recorded. Other covers Western Default, Repetition, and Incomplete Response as primary labels.}
    \label{fig:verdicts}
\end{figure}

\subsection{Error Analysis}
\label{sec:error_analysis}
Thirteen error categories were applied during human evaluation: Hallucination, Shallow Description, Model Breakdown, Repetition, Western Default, Nonsensical Response, Empty Response, Incomplete Response, Culturally Accurate, Pan-African Generalisation, Ethnolinguistic Substitution, Wrong Language, and Contradicts Prompt. A single response often receives multiple labels. The two smallest models collapse entirely under Igbo-language input. LLaMA 3 8B Lite averages 3.034 on Type~1 but drops to 0.000 on Type~2 and 0.102 on Type~3. LLaMA 3.1 8B follows the same pattern, falling from 2.983 (Type~1) to 0.153 (Type~3), a 94.9\% drop. In Type~1, Hallucination is the most frequent error, followed by Shallow Description and Western Default. Across Type~2 and Type~3, Western Default declines as models can no longer produce coherent output, while Model Breakdown and Repetition rise sharply, from 17 and 0 labelled responses in Type~1 to 158 and 124 in Type~3 (see Figure \ref{fig:repetition} in the Appendix for examples of repetition-loop failures). Figure~\ref{fig:verdicts} shows the primary verdict for each answer: in Type~1, 93\% of answers are primarily hallucinations and 4\% are correct; under Igbo input, 27\% (Type~2) and 33\% (Type~3) of answers are primarily model breakdowns, and no prompt type exceeds 7\% correct. To illustrate how the distractor taxonomy targets these failure modes, consider the communal proverb \textit{``E kechaa n’obi eke na mkpuke''}, which affirms that resources are shared first on the basis of the extended family and then on the basis of the nuclear family. Its Western Default distractor reads: \textit{``Resources must be strictly divided into equal individual portions, disregarding family hierarchies, to promote absolute egalitarianism.''} This option replaces the Igbo family hierarchy with Western individualism, so a model that selects it reveals a specific misunderstanding rather than a generic error. Pan-African Generalisation distractors work in the same way, by replacing specific Igbo practices with broad continental stereotypes that strip away their localized meaning.

\subsection{Cross-Model Synthesis}
\label{sec:synthesis}
The central empirical finding is not low MCQ accuracy but the magnitude of the gap between MCQ performance and open-ended cultural generation. Figure~\ref{fig:gap} places the two formats side by side on the same 59 questions in the zero-shot condition. GPT-4o mini selects the correct option for 87.0\% of MCQ items but produces a fully correct explanation (EF $=2$) for only 2.3\% of open-ended answers. Even Gemini 2.0 Flash, the strongest model, reaches 97.2\% on MCQs but only 26.0\% fully correct explanations, and no other model exceeds 4.5\%. Over the full benchmark, GPT-4o mini achieves 93.2\% overall MCQ accuracy while producing Western Default explanations in open-ended tasks. AfriqueQwen 14B scores 88.2\% on MCQs yet generates invented social labels and incorrect ritual meanings. LLaMA 3 8B Lite reaches 71.8\% MCQ accuracy but shows repeated generation failure under Igbo prompting. GPT-OSS 120B shows the same pattern: it ranks third in MCQ accuracy at 89.9\% over the full benchmark, yet on the 59 human-evaluated questions only 3.4\% of its open-ended answers are fully correct. This confirms that multiple-choice formats reward pattern matching, not genuine cultural reasoning. Model scale does not determine cultural alignment. GPT-OSS 120B is the largest open model, but its high MCQ accuracy does not carry over to open-ended cultural reasoning. LLaMA 3.3 70B reaches 99.0\% on Type~1 MCQs but collapses to 46.1\% on Type~3. This 52.9 percentage point drop is the largest degradation observed and is consistent with the ``fluent but foreign'' phenomenon described by \citet{Agarwal2025FluentBF}. Our discovery that models can sound fluent without understanding culture also mirrors the findings in XCR-Bench \citep{xcrbench2024}. Surface fluency does not guarantee deep cultural grounding. The two best models on both evaluation tracks (Gemini 2.0 Flash and GPT-4o mini) are proprietary models accessible only through paid commercial APIs. The communities most harmed by cultural misrepresentation are the same low-resource communities for whom such access is prohibitive. This dissociation between knowledge (EF/RR) and fluency (LQ/PP) reinforces our central claim: cultural misalignment in current LLMs is not primarily a multilingual generation problem. Even prompted entirely in English, most models fail to produce culturally grounded reasoning at rates statistically indistinguishable from their Igbo-prompted outputs; Gemini 2.0 Flash is the only model whose cultural knowledge scores improve in Igbo. The models have not learned Igbo culture in any language.

\section{Conclusion}
We introduced CAB-Igbo, a benchmark for evaluating cultural alignment in Igbo contexts. The dataset comprises 296 questions across two regional states and six cultural domains. All questions are drawn from authentic community sources and validated by native speakers. Evaluations on 8 LLMs reveal significant performance gaps. Open-weight and proprietary models both struggle with Igbo cultural contexts. They frequently default to Western norms or Pan-African generalizations. The gap between MCQ accuracy and open-ended reasoning is severe. While models can recognize patterns, they consistently struggle to generate culturally grounded explanations. This highlights the urgent need for culturally aware models built directly from community knowledge, rather than relying on translated Western concepts or superficial fluency. Addressing this gap is not just an academic exercise. It highlights the severe danger of confusing language fluency with cultural understanding. As AI expands globally, deploying models without cultural grounding risks concrete harm in education, public services, and safety. Cultural alignment is not a feature; it is the foundation for safe and trustworthy AI.
\section*{Limitations}
\textbf{Statistical Significance \& Positional Bias:} Our significance tests treat items as independent and apply no correction for multiple comparisons across models and conditions, so results near the 0.05 threshold should be read with caution. Furthermore, evaluating with only two answer permutations is not robust enough to definitively rule out positional bias. Finally, we observed extremely high Type-1 MCQ accuracies near ceiling for frontier models. Future iterations should incorporate item-level difficulty analysis (e.g., Item Response Theory) to better calibrate English-prompted difficulty.\\
\\
\textbf{Two Evaluation Runs for the GPT-OSS Models:} The MCQ results for GPT-OSS 120B combine two runs: the first run, and a second run that queried again only the 5,640 items whose first-run response was empty because the model's reasoning exceeded the 300-token limit. The second run used a 4,000-token limit and otherwise identical settings; first-run responses that contained an answer were kept unchanged. GPT-OSS 20B could not be re-queried on the same service, so 11,826 of its 19,536 MCQ responses (60.5\%) remain empty and are scored as incorrect. On the items it did answer (original ordering), GPT-OSS 20B selected the correct option 97.0\% of the time, so its reported MCQ accuracy mainly reflects empty output rather than cultural knowledge, and we do not draw conclusions about cultural competence from it. The open-ended responses of both models come from the first run, with empty answers scored 0 as specified by the rubric.\\
\\
\textbf{Human Evaluation Constraints:} Due to severe resource constraints, the open-ended evaluation was conducted by a single native-speaker expert linguist. The absence of inter-human agreement clouds our conclusions regarding construct-level score distributions, especially for Epistemic Fidelity and Representational Richness. In future work, we will conduct a dual-human calibration study on 100-200 items to cleanly separate LLM judge failures from single-rater biases.\\
\\
\textbf{Decoding Strategies \& Prompting Gaps:} We relied exclusively on zero-temperature greedy decoding to enforce deterministic outputs. This may have disproportionately disadvantaged some models on the generative short-answer questions (SAQs). Robustness to varying decoding temperatures, few-shot prompting, or retrieval-augmented generation (RAG) was not probed. Additionally, because the open-ended generations were not evaluated under persona and state conditions, the interaction between demographic prompting and generative cultural fidelity remains an under-explored gap.\\
\\
\textbf{The Influence of Standardisation and Dialects on Cultural Reasoning:} To ensure a stable baseline, we utilized Igbo Izugbe, or Standard Central Igbo. This standardisation is heavily linked to our Imo State persona. However, it introduces a limitation: models that perform well on standard Igbo often struggle with dialectal or local knowledge. We included Ebonyi State to test intra-Igbo dialectal variation, but certain cultural concepts from other regions remain underrepresented. Preliminary evidence suggests that dialectal coverage materially changes Epistemic Fidelity and Representational Richness scores, making expansion beyond Imo and Ebonyi a critical next step.\\
\\
\textbf{Coverage of Regions and Evaluation Constraints}
Our study covers six foundational domains of Igbo cultural life. However, cultural knowledge is not static. It varies widely across communities. Due to severe resource, API, and GPU constraints, we restricted our human evaluation phase. We sampled only 20\% of the full benchmark. This evaluation was conducted exclusively under zero-shot conditions. We did not assess the open-ended persona responses. Additionally, a single native speaker evaluator conducted the manual evaluation. This introduces potential subjectivity, despite the rigorous source-grounded rubric. Future work should expand the benchmark. It should include more underrepresented dialects, contemporary cultural practices, and broader native-speaker adjudication.

\section*{Ethics and Data Availability}
CAB-Igbo will be publicly released with open licenses, complete annotation guidelines, and evaluation scripts to enable external audits and future research. We maintained strict privacy and informed consent throughout the dataset creation. All native community contributors were fairly compensated for their expertise.

\section*{GenAI Usage Disclosure}
Large language models (GPT-4o mini, Gemini 2.0 Flash, LLaMA 3.3 70B, LLaMA 3.1 8B, 
LLaMA 3 8B Lite, GPT-OSS 120B, GPT-OSS 20B, and AfriqueQwen 14B) are the objects of 
study in this paper and were used exclusively to generate the evaluated responses 
described in Section~\ref{sec:experimental_setup}. Additionally, Claude Sonnet 5 
(Anthropic) was used to assist with code debugging, copy editing, and LaTeX 
formatting, under full author review and verification. No GenAI tool was used to 
generate, fabricate, or alter any cultural content, dataset item, or human 
evaluation score reported in this work.

\bibliography{custom}
\appendix
\raggedbottom

\section{Detailed Results}
\label{app:results}

\subsection{Multiple-choice accuracy}
Tables~\ref{tab:prompt_type} to~\ref{tab:persona_detail} report MCQ accuracy in the original answer ordering.

\begin{table}[H]
\centering\small
\begin{tabular}{@{}lrrr@{}}
\toprule
\textbf{Model} & \textbf{Type 1} & \textbf{Type 2} & \textbf{Type 3} \\
\midrule
Gemini 2.0 Flash    & 99.2 & 97.4 & 98.1 \\
GPT-4o mini         & 98.2 & 90.2 & 91.1 \\
AfriqueQwen 14B     & 99.5 & 90.7 & 74.4 \\
LLaMA 3.3 70B       & 99.0 & 91.6 & 46.1 \\
LLaMA 3 8B Lite     & 96.8 & 53.7 & 64.8 \\
LLaMA 3.1 8B        & 96.3 & 60.0 & 51.5 \\
GPT-OSS 120B        & 96.4 & 86.7 & 86.6 \\
GPT-OSS 20B         & 73.0 & 12.9 & 28.7 \\
\bottomrule
\end{tabular}
\caption{MCQ accuracy (\%) by linguistic prompt type, pooled over the baseline and all persona conditions.}
\label{tab:prompt_type}
\end{table}

\begin{table}[H]
\centering\small
\begin{tabular}{@{}lrrr@{}}
\toprule
\textbf{Model} & \textbf{Baseline} & \textbf{Imo} & \textbf{Ebonyi} \\
\midrule
Gemini 2.0 Flash    & 98.2 & 98.2 & 98.3 \\
GPT-4o mini         & 91.7 & 93.5 & 93.2 \\
AfriqueQwen 14B     & 93.2 & 87.7 & 87.8 \\
LLaMA 3.3 70B       & 81.5 & 78.7 & 78.5 \\
LLaMA 3 8B Lite     & 71.6 & 71.6 & 71.9 \\
LLaMA 3.1 8B        & 59.8 & 70.1 & 70.3 \\
GPT-OSS 120B        & 89.6 & 90.0 & 89.9 \\
GPT-OSS 20B         & 45.9 & 37.7 & 37.1 \\
\bottomrule
\end{tabular}
\caption{MCQ accuracy (\%) for the zero-shot baseline and for the Imo and Ebonyi conditions, each pooled over the five personas.}
\label{tab:state}
\end{table}

\begin{table}[H]
\centering\small
\begin{tabular}{@{}lrrr@{}}
\toprule
\textbf{Model} & \textbf{Baseline} & \textbf{All personas} & \textbf{Delta} \\
\midrule
Gemini 2.0 Flash    & 98.2 & 98.3 & $+$0.1 \\
GPT-4o mini         & 91.7 & 93.3 & $+$1.6 \\
AfriqueQwen 14B     & 93.2 & 87.7 & $-$5.5 \\
LLaMA 3.3 70B       & 81.5 & 78.6 & $-$2.9 \\
LLaMA 3 8B Lite     & 71.6 & 71.8 & $+$0.2 \\
LLaMA 3.1 8B        & 59.8 & 70.2 & $+$10.4 \\
GPT-OSS 120B        & 89.6 & 89.9 & $+$0.3 \\
GPT-OSS 20B         & 45.9 & 37.4 & $-$8.5 \\
\bottomrule
\end{tabular}
\caption{MCQ accuracy (\%) under the zero-shot baseline and pooled over all ten persona conditions; Delta is in percentage points.}
\label{tab:persona_baseline}
\end{table}

\begin{table}[H]
\centering\small
\setlength{\tabcolsep}{4pt}
\begin{tabular}{@{}lrrrrrr@{}}
\toprule
\textbf{Model} & \textbf{BL} & \textbf{YTB} & \textbf{MW} & \textbf{MM} & \textbf{WW} & \textbf{Prof} \\
\midrule
Gemini 2.0 Flash  & 98.2 & 98.3 & 98.5 & 98.1 & 98.1 & 98.3 \\
GPT-4o mini       & 91.7 & 93.3 & 92.8 & 93.2 & 93.6 & 93.6 \\
AfriqueQwen 14B   & 93.2 & 88.5 & 87.5 & 87.7 & 87.8 & 87.1 \\
LLaMA 3.3 70B     & 81.5 & 80.0 & 78.6 & 78.6 & 77.4 & 78.4 \\
LLaMA 3 8B Lite   & 71.6 & 71.7 & 72.6 & 71.8 & 71.7 & 71.1 \\
LLaMA 3.1 8B      & 59.8 & 68.5 & 70.8 & 71.3 & 67.8 & 72.6 \\
GPT-OSS 120B      & 89.6 & 90.1 & 89.9 & 89.4 & 90.3 & 90.0 \\
GPT-OSS 20B       & 45.9 & 38.1 & 35.9 & 36.3 & 37.8 & 38.9 \\
\bottomrule
\end{tabular}
\caption{MCQ accuracy (\%) by persona, pooled over Imo and Ebonyi. BL = zero-shot baseline; YTB = Young Teenage Boy; MW = Married Woman; MM = Married Man; WW = Widowed Woman; Prof = Professor.}
\label{tab:persona_detail}
\end{table}

\subsection{Significance tests}

\begin{table}[H]
\centering\small
\resizebox{\columnwidth}{!}{%
\begin{tabular}{@{}lrlrlc@{}}
\toprule
 & \multicolumn{2}{c}{\textbf{EN $\rightarrow$ IG}} & \multicolumn{2}{c}{\textbf{Position}} & \textbf{Persona} \\
\cmidrule(lr){2-3}\cmidrule(lr){4-5}
\textbf{Model} & $\chi^2$ & $p$ & $\chi^2$ & $p$ & \textbf{sig.} \\
\midrule
AfriqueQwen 14B  & 31.03  & $<0.001$ & 0.12  & 0.731    & 10/10 \\
Gemini 2.0 Flash & 5.14   & 0.023    & 5.26  & 0.022    & 0/10 \\
GPT-4o mini      & 24.74  & $<0.001$ & 16.60 & $<0.001$ & 7/10 \\
GPT-OSS 120B     & 25.29  & $<0.001$ & 0.13  & 0.716    & 0/10 \\
GPT-OSS 20B      & 164.28 & $<0.001$ & 0.67  & 0.413    & 10/10 \\
LLaMA 3.1 8B     & 113.39 & $<0.001$ & 11.52 & $<0.001$ & 10/10 \\
LLaMA 3.3 70B    & 12.19  & $<0.001$ & 4.34  & 0.037    & 9/10 \\
LLaMA 3 8B Lite  & 113.39 & $<0.001$ & 3.36  & 0.067    & 0/10 \\
\bottomrule
\end{tabular}}
\caption{McNemar's tests ($\chi^2$ with Yates' continuity correction) on paired MCQ outcomes for the zero-shot baseline items. EN $\rightarrow$ IG: Type~1 vs Type~2, original ordering (296 pairs). Position: original vs shuffled ordering (888 question and prompt type pairs). Persona: number of the ten persona conditions that differ significantly ($p<0.05$) from the baseline (888 pairs each, original ordering).}
\label{tab:mcnemar_results}
\end{table}

\subsection{Human evaluation}

\begin{table}[H]
\centering\small
\begin{tabular}{@{}lrrr@{}}
\toprule
\textbf{Model} & \textbf{Type 1} & \textbf{Type 2} & \textbf{Type 3} \\
\midrule
Gemini 2.0 Flash   & 3.559 & 4.441 & 4.949 \\
GPT-4o mini        & 3.271 & 3.169 & 3.000 \\
GPT-OSS 120B       & 3.102 & 3.153 & 3.017 \\
LLaMA 3.3 70B      & 3.203 & 3.119 & 2.797 \\
AfriqueQwen 14B    & 3.237 & 2.983 & 2.356 \\
GPT-OSS 20B        & 2.729 & 1.915 & 1.741 \\
LLaMA 3.1 8B       & 2.983 & 0.678 & 0.153 \\
LLaMA 3 8B Lite    & 3.034 & 0.000 & 0.102 \\
\midrule
All models (mean)  & 3.140 & 2.432 & 2.265 \\
\bottomrule
\end{tabular}
\caption{Mean total human evaluation score (out of 8) by model and prompt type.}
\label{tab:type_degradation}
\end{table}

\begin{table}[H]
\centering\small
\begin{tabular}{@{}lrrrr@{}}
\toprule
\textbf{Model} & \textbf{EF} & \textbf{RR} & \textbf{LQ} & \textbf{PP} \\
\midrule
Gemini 2.0 Flash   & 0.531 & 0.531 & 1.994 & 1.260 \\
GPT-4o mini        & 0.062 & 0.062 & 2.000 & 1.023 \\
GPT-OSS 120B       & 0.074 & 0.074 & 1.938 & 1.006 \\
LLaMA 3.3 70B      & 0.062 & 0.062 & 1.932 & 0.983 \\
AfriqueQwen 14B    & 0.102 & 0.102 & 1.740 & 0.915 \\
GPT-OSS 20B        & 0.040 & 0.040 & 1.351 & 0.698 \\
LLaMA 3.1 8B       & 0.034 & 0.034 & 0.803 & 0.401 \\
LLaMA 3 8B Lite    & 0.006 & 0.006 & 0.689 & 0.345 \\
\bottomrule
\end{tabular}
\caption{Mean human evaluation score per construct (each 0 to 2), averaged over the three prompt types.}
\label{tab:human_overall}
\end{table}

\begin{table}[H]
\centering\small
\setlength{\tabcolsep}{4.5pt}
\begin{tabular}{@{}lrrrr@{}}
\toprule
\textbf{Model} & \textbf{$\Delta$EF} & \textbf{$\Delta$RR} & \textbf{$\Delta$LQ} & \textbf{$\Delta$PP} \\
\midrule
AfriqueQwen 14B  & $-0.03$     & $-0.03$     & $0.24^{*}$  & $0.08$      \\
Gemini 2.0 Flash & $-0.34^{*}$ & $-0.34^{*}$ & $-0.02$     & $-0.19^{*}$ \\
GPT-4o mini      & $0.05$      & $0.05$      & $0.00$      & $0.00$      \\
GPT-OSS 120B     & $-0.08$     & $-0.08$     & $0.10$      & $0.02$      \\
GPT-OSS 20B      & $0.05$      & $0.05$      & $0.46^{*}$  & $0.25^{*}$  \\
LLaMA 3.1 8B     & $0.10^{*}$  & $0.10^{*}$  & $1.36^{*}$  & $0.75^{*}$  \\
LLaMA 3.3 70B    & $-0.02$     & $-0.02$     & $0.07^{*}$  & $0.05$      \\
LLaMA 3 8B Lite  & $0.02$      & $0.02$      & $2.00^{*}$  & $1.00^{*}$  \\
\bottomrule
\end{tabular}
\caption{Mean within-item score difference (Type~1 minus Type~2) in human evaluation. Positive values are drops under Igbo prompting; negative values are gains. Asterisks mark differences whose bootstrap 95\% confidence interval (10,000 resamples) excludes zero. Each model has 59 paired items.}
\label{tab:bootstrap_results}
\end{table}

\subsection{LLM-as-a-judge}

\begin{table}[H]
\centering\small
\setlength{\tabcolsep}{4.5pt}
\begin{tabular}{@{}lrrrrr@{}}
\toprule
\textbf{Model} & $n$ & \textbf{EF} & \textbf{RR} & \textbf{LQ} & \textbf{PP} \\
\midrule
GPT-OSS 120B    & 177 & 1.113 & 1.023 & 1.989 & 1.955 \\
GPT-4o mini     & 177 & 0.672 & 0.582 & 2.000 & 2.000 \\
LLaMA 3.3 70B   & 177 & 0.599 & 0.531 & 1.938 & 1.927 \\
AfriqueQwen 14B & 177 & 0.644 & 0.542 & 1.751 & 1.734 \\
GPT-OSS 20B     & 176 & 0.619 & 0.597 & 1.460 & 1.449 \\
LLaMA 3.1 8B    & 177 & 0.226 & 0.175 & 1.011 & 1.023 \\
LLaMA 3 8B Lite & 177 & 0.141 & 0.119 & 0.910 & 0.831 \\
\bottomrule
\end{tabular}
\caption{Mean LLM-as-a-judge scores by model, ordered by total, over the responses matched to human scores (59 questions across three prompt types); $n$ is the number of matched responses. Empty responses are scored 0, as in the human rubric.}
\label{tab:judge_scores}
\end{table}

\begin{table}[H]
\centering\small
\setlength{\tabcolsep}{4.2pt}
\begin{tabular}{@{}lrrrrrr@{}}
\toprule
 & $\kappa_w$ & $\rho$ & \textbf{Exact} & \textbf{Adj.} & \textbf{Human} & \textbf{Judge} \\
\midrule
EF & 0.112 & 0.276 & 59.8 & 87.5 & 0.054 & 0.574 \\
RR & 0.100 & 0.240 & 59.9 & 92.2 & 0.054 & 0.510 \\
LQ & 0.833 & 0.885 & 90.9 & 97.3 & 1.494 & 1.580 \\
PP & 0.261 & 0.883 & 22.1 & 98.1 & 0.767 & 1.560 \\
\bottomrule
\end{tabular}
\caption{Agreement between the human annotator and the Gemini 2.0 Flash judge over 1,238 matched responses. $\kappa_w$ = weighted Cohen's kappa; $\rho$ = Spearman rank correlation; Exact and Adj.\ = percentage of pairs that agree exactly or differ by at most one point; Human and Judge = mean scores.}
\label{tab:ira}
\end{table}

\subsection{Repetition failures}
Figure~\ref{fig:repetition} shows verbatim excerpts of the repetition loops described in Section~\ref{sec:error_analysis}.

\input{fig_repetition.tex}

\section{Dataset Statement for CAB-Igbo}
\label{app:dataset}

\subsection{Creation}
CAB-Igbo is built entirely from scratch, without relying on web-scraped text. All data is manually created and validated by native speakers from two Igbo-speaking states. The dataset contains 296 questions in Igbo and English. Questions were designed and reviewed by a native Igbo speaker and an Igbo linguist.

\subsection{Cultural domains}
CAB-Igbo evaluates six domains of Igbo cultural life. Table~\ref{tab:domains} gives the number of questions in each.

\begin{table}[H]
\centering\small
\begin{tabular}{@{}lr@{}}
\toprule
\textbf{Domain} & \textbf{Questions} \\
\midrule
Food and Cuisine             & 73 \\
Festivals and Religion       & 67 \\
Family and Kinship           & 58 \\
Proverbs and Oral Tradition  & 38 \\
Social Customs and Etiquette & 35 \\
Funerals and Burial          & 25 \\
\midrule
Total                        & 296 \\
\bottomrule
\end{tabular}
\caption{Number of questions per cultural domain.}
\label{tab:domains}
\end{table}

\subsection{Question design}
Every question must:
\begin{itemize}
    \item test authentic Igbo cultural knowledge;
    \item contain one clear and defensible correct answer;
    \item reflect cultural practices among Igbo communities.
\end{itemize}
Questions test cultural reasoning rather than factual recall alone. Preference was given to questions that require an understanding of cultural logic and meaning, knowledge of ritual and social context, and recognition of appropriate behaviour within Igbo society. For example, some questions ask why a child may be buried without mourning rites; the correct answer requires an understanding of the \textit{ogbanje} belief system rather than simple memorisation.

Questions use direct and unambiguous language: they avoid vague wording, specify traditional or modern practice where necessary, and use precise cultural terminology.

\subsection{Linguistic grid and question formats}
Each question is evaluated in three language conditions: Type~1 (English instruction, English question), Type~2 (English instruction, Igbo question), and Type~3 (Igbo instruction, Igbo question).

Each multiple-choice question contains one correct answer and three diagnostic distractors. Distractors represent different forms of cultural misunderstanding, so the benchmark identifies how models fail, not only whether they fail. Each multiple-choice item has a matching open-ended version, which requires the model to explain the cultural reasoning in two to four sentences without answer choices. Open-ended responses are evaluated through human annotation using the rubric in Appendix~\ref{app:annotation}.

\subsection{Distractor taxonomy}
Each distractor belongs to exactly one of four categories, assigned during dataset construction (Table~\ref{tab:distractors}).

\begin{table}[H]
\centering\small
\begin{tabular}{@{}L{0.27\columnwidth}L{0.49\columnwidth}r@{}}
\toprule
\textbf{Category} & \textbf{Definition} & \textbf{Count} \\
\midrule
Western Default & Interprets Igbo culture through Western assumptions such as individualism or nuclear family structures & 292 \\
\addlinespace
Shallow Description & Recognises surface patterns without deeper cultural understanding & 256 \\
\addlinespace
Ethnolinguistic Substitution & Replaces Igbo practices with Yoruba or Hausa cultural forms & 174 \\
\addlinespace
Pan-African Generalisation & Uses broad African stereotypes that erase Igbo-specific meaning & 166 \\
\midrule
Total & & 888 \\
\bottomrule
\end{tabular}
\caption{Distractor categories and their frequency across the 888 distractors (three per question).}
\label{tab:distractors}
\end{table}

\subsection{Bilingual and orthographic requirements}
Every question appears in parallel Igbo and English versions, and both versions must test the same knowledge with the same meaning. The Igbo version must read naturally and use appropriate Igbo terminology for cultural concepts. All Igbo text follows standard orthography, includes tone marks where necessary, and keeps spelling consistent throughout the dataset.

\subsection{Item structure}
Each item records the question number, the cultural domain, the question text in Igbo and English, four answer options (A to D) in both languages, the correct answer label, and the distractor category of each incorrect option.

\subsection{Quality criteria}
Before inclusion, each question must satisfy the following conditions:
\begin{enumerate}
    \item The correct answer reflects authentic Igbo practice.
    \item The question and options are unambiguous.
    \item Distractors are plausible but incorrect.
    \item Each distractor reveals a specific misunderstanding.
    \item The Igbo and English versions are semantically equivalent.
    \item The question tests unique knowledge within the dataset.
\end{enumerate}

\subsection{Review process}
Items passed through three stages.
\begin{description}
    \item[Question creation.] Questions were drafted independently, with balanced domain coverage, and each distractor was assigned a category label.
    \item[Joint review.] Each question was reviewed collaboratively to verify cultural accuracy, check translation equivalence, resolve ambiguity, and confirm the distractor categories.
    \item[Quality control.] Duplicate and ambiguous questions were removed, the domain distribution was checked, and formatting was finalised.
\end{description}

\subsection{Native speaker validation}
Native Igbo speakers reviewed all items using a participatory evaluation process, checking cultural accuracy, linguistic quality, and contextual fit. Items without agreement were revised or removed. This process grounds the benchmark in lived cultural expertise rather than external interpretation or synthetic generation.

\subsection{Common pitfalls}
Dataset creators were asked to avoid:
\begin{itemize}
    \item regional variation that produces multiple valid answers;
    \item questions with more than one defensible answer;
    \item superficial questions that test recall instead of understanding;
    \item translation errors or unnatural Igbo phrasing;
    \item weak distractors that are obviously incorrect;
    \item unnecessary gender stereotypes unless culturally required.
\end{itemize}

\subsection{Documentation}
The development process records the rationale for each question design, removed questions and the reasons for removal, revisions made during review, distractor category decisions, and domain distribution statistics.

\section{Annotator Guidelines}
\label{app:annotation}

\subsection{Background}
The CAB-Igbo benchmark evaluates how AI models represent and apply Igbo cultural knowledge. In this study, trained annotators assess model-generated responses to cultural questions using a standardised rubric. The evaluation has two objectives: first, to validate the automated LLM-as-a-judge system, and second, to compare model performance across linguistic prompt types. Model identities remain hidden throughout the evaluation; annotators see only anonymised labels such as \textit{Model A} and \textit{Model B}.

\subsection{Evaluation setup}
Annotators evaluate 20\% of the CAB-Igbo dataset, and each question features responses from eight models. All responses are generated under the zero-shot baseline condition (no persona), in all three prompt types (Type~1: English instruction, English question; Type~2: English instruction, Igbo question; Type~3: Igbo instruction, Igbo question). The questions cover the six domains listed in Table~\ref{tab:domains}.

\subsection{Scoring constructs}
Each response is scored on four constructs using a 0 to 2 integer scale (Table~\ref{tab:rubric}). For Igbo responses (Type~2 and Type~3), annotators assess Igbo grammar and morphology directly under Linguistic Quality. The total response score is the sum of the four constructs, $\mathrm{EF}+\mathrm{RR}+\mathrm{LQ}+\mathrm{PP}$, and ranges from 0 to 8.

\begin{table}[H]
\centering\small
\begin{tabular}{@{}c p{0.80\columnwidth}@{}}
\toprule
\multicolumn{2}{@{}p{0.92\columnwidth}@{}}{\textbf{Epistemic Fidelity (EF):} factual accuracy and cultural reliability} \\
\midrule
2 & Completely accurate and culturally specific \\
1 & Partially accurate or surface-level \\
0 & Completely incorrect, fabricated, or blank \\
\midrule
\multicolumn{2}{@{}p{0.92\columnwidth}@{}}{\textbf{Representational Richness (RR):} depth and nuance} \\
\midrule
2 & Rich, nuanced, and culturally grounded \\
1 & Basic or vague cultural detail \\
0 & No meaningful cultural content, or blank \\
\midrule
\multicolumn{2}{@{}p{0.92\columnwidth}@{}}{\textbf{Linguistic Quality (LQ):} fluency, grammar, and language use} \\
\midrule
2 & Fluent and grammatically correct \\
1 & Understandable despite noticeable errors \\
0 & Largely unintelligible, wrong language, or blank \\
\midrule
\multicolumn{2}{@{}p{0.92\columnwidth}@{}}{\textbf{Pragmatic Proficiency (PP):} contextual appropriateness} \\
\midrule
2 & Fully appropriate and context-aware \\
1 & Partially appropriate or with minor inconsistencies \\
0 & Completely misaligned, off-topic, or blank \\
\bottomrule
\end{tabular}
\caption{Human evaluation rubric.}
\label{tab:rubric}
\end{table}

\subsection{Annotation procedure}
Annotators use a custom web-based evaluation interface and complete the following steps.
\begin{enumerate}
    \item Enter the authorised \textbf{Annotator ID} to log in and load the assigned evaluation set.
    \item For each grouped question block:
    \begin{enumerate}
        \item Read the cultural question carefully, reviewing both the English and Igbo versions to ensure a full understanding of the cultural context.
        \item Note the prompt type (Type~1, Type~2, or Type~3) displayed for the block.
        \item Read the corresponding model-generated response.
        \item Use the interface dropdowns to assign an integer score from 0 to 2 for EF, RR, LQ, and PP.
        \item Add any necessary clarification in the optional \textbf{Comments} box.
    \end{enumerate}
    \item Click \textbf{Submit} at the end of each block to register the scores.
    \item Click \textbf{Download CSV} periodically to back up progress. Once all questions are evaluated, perform a final download and submit the file to the lead researchers.
\end{enumerate}

\subsection{Evaluation rules}
\begin{itemize}
    \item Evaluate each response independently.
    \item Responses marked \texttt{[No response found]} or left completely blank receive a score of 0 for all constructs.
    \item Assess Igbo fluency under Linguistic Quality for Type~2 and Type~3 tasks.
    \item Do not discuss scores with other annotators before completing the evaluation.
    \item If uncertain between two scores, assign the lower score and add a brief comment.
\end{itemize}

\input{appendix_generated.tex}

\end{document}

%% file: fig_samples.tex
{\sffamily\small\renewcommand{\arraystretch}{1.35}
\begin{tabular}{@{}L{0.17\textwidth}@{\hspace{0.03\textwidth}}L{0.385\textwidth}@{\hspace{0.03\textwidth}}L{0.385\textwidth}@{}}
\specialrule{1.4pt}{0pt}{4pt}
\textbf{\footnotesize CULTURAL DOMAIN} & \textbf{\footnotesize ENGLISH PROMPT} & \textcolor{igbogreen}{\textbf{\footnotesize IGBO PROMPT}} \\
\specialrule{1.0pt}{4pt}{6pt}
\textbf{Proverbs and Oral Tradition} & \textcolor{englishgray}{What practical life lesson is contained in the proverb, \textquotedbl{}It is the height that a child\textquotesingle{}s hand gets to that he hangs his bag\textquotedbl{}?} & \textcolor{igbogreen}{\textbf{Kedu nkuzi ndụ bara uru dị n\textquotesingle{}ime ilu a: \textquotedbl{}Ebe eka nwata ru ke o kọnto ekpa\textquotedbl{}?}} \\
\specialrule{0.4pt}{6pt}{6pt}
\textbf{Family and Kinship} & \textcolor{englishgray}{In Igbo kinship and hospitality rituals, how is an Ọjị Igbo (kola nut) that breaks into exactly five cotyledons interpreted by the family?} & \textcolor{igbogreen}{\textbf{N\textquotesingle{}omenala mmekọrịta ezinụlọ na ile ọbịa ndị Igbo, kedu ka ezinụlọ si akọwa Ọjị Igbo gbara kpọmkwem ibe ise?}} \\
\specialrule{0.4pt}{6pt}{6pt}
\textbf{Funerals and Burial} & \textcolor{englishgray}{In traditional Igbo cosmology, which of the following criteria defines the attainment of a \textquotedbl{}good death\textquotedbl{} (Ọnwụ chi)?} & \textcolor{igbogreen}{\textbf{N\textquotesingle{}ọdịnala na amamihe okike ndị Igbo, olee n\textquotesingle{}ime ọnọdụ ndị a na-akọwa inweta \textquotedbl{}ezi ọnwụ\textquotedbl{} (Ọnwụ chi)?}} \\
\specialrule{1.4pt}{6pt}{0pt}
\end{tabular}}

%% file: fig_repetition.tex
\begin{figure*}[t]
\centering\sffamily\small\renewcommand{\arraystretch}{1.2}
\begin{tabular}{@{}L{0.47\textwidth}@{\hspace{0.06\textwidth}}L{0.47\textwidth}@{}}
\specialrule{1.2pt}{0pt}{6pt}
\textbf{LLaMA 3 8B Lite}\hfill{\footnotesize\color{statgray}Type 2 \textperiodcentered{} question 1}\par\smallskip \loopbar{0.994}\hspace{6pt}{\footnotesize\textbf{99\%} of 167 words \textperiodcentered{} 166 repeats}\par\smallskip {\footnotesize\setlength{\fboxsep}{1pt}Onye \loopword{agha} \loopword{agha} \loopword{agha} \loopword{agha} \loopword{agha} \loopword{agha} \loopword{agha} \loopword{agha} \loopword{agha} \loopword{agha} \loopword{agha} \loopword{agha} \loopword{agha} \loopword{agha} \loopword{agha} \loopword{agha} \loopword{agha} \loopword{agha} \loopword{agha} \loopword{agha} \loopword{agha} \loopword{agha} \loopword{agha} \loopword{agha} \loopword{agha} \loopword{agha} \loopword{agha} \loopword{agha} \loopword{agha} \dots} & \textbf{LLaMA 3.1 8B}\hfill{\footnotesize\color{statgray}Type 2 \textperiodcentered{} question 1}\par\smallskip \loopbar{0.709}\hspace{6pt}{\footnotesize\textbf{71\%} of 86 words \textperiodcentered{} 61 repeats}\par\smallskip {\footnotesize\setlength{\fboxsep}{1pt}\dots{} dị ka Ọnwụma \loopword{Ọjọọ,} Ọnwụma \loopword{Ọjọọ} \loopword{Ọjọọ,} Ọnwụma \loopword{Ọjọọ} \loopword{Ọjọọ} \loopword{Ọjọọ,} Ọnwụma \loopword{Ọjọọ} \loopword{Ọjọọ} \loopword{Ọjọọ} \loopword{Ọjọọ,} Ọnwụma \loopword{Ọjọọ} \loopword{Ọjọọ} \loopword{Ọjọọ} \loopword{Ọjọọ} \loopword{Ọjọọ,} Ọnwụma \loopword{Ọjọọ} \loopword{Ọjọọ} \loopword{Ọjọọ} \loopword{Ọjọọ} \loopword{Ọjọọ} \loopword{Ọjọọ,} Ọnwụma \dots} \\
\specialrule{0.4pt}{7pt}{7pt}
\textbf{GPT-OSS 20B}\hfill{\footnotesize\color{statgray}Type 2 \textperiodcentered{} question 5}\par\smallskip \loopbar{0.364}\hspace{6pt}{\footnotesize\textbf{36\%} of 77 words \textperiodcentered{} 4 repeats}\par\smallskip {\footnotesize\setlength{\fboxsep}{1pt}\dots{} nwa ọhụrụ nwere \loopword{àgwà} \loopword{ma} \loopword{ọ} \loopword{bụ} \loopword{àgwà} \loopword{dị} \loopword{iche,} dịka ịma mma ma ọ bụ obi ike, nke yiri nke nna nna ahụ. Ọtụtụ mgbe, a na-ahụkwa akara anụ ahụ dịka \loopword{àgwà} \loopword{ma} \loopword{ọ} \loopword{bụ} \loopword{àgwà} \loopword{dị} \loopword{iche,} \dots} & \textbf{AfriqueQwen 14B}\hfill{\footnotesize\color{statgray}Type 3 \textperiodcentered{} question 5}\par\smallskip \loopbar{0.505}\hspace{6pt}{\footnotesize\textbf{51\%} of 182 words \textperiodcentered{} 4 repeats}\par\smallskip {\footnotesize\setlength{\fboxsep}{1pt}\dots{} na-akwado nkwupụta a. \loopword{Ụfọdụ} \loopword{ndị} \loopword{na-akọ} \loopword{akụkọ} \loopword{banyere} \loopword{ihe} \loopword{ndị} \loopword{dị} \loopword{ka} \loopword{ọdịdị} \loopword{anụ} \loopword{ahụ} \loopword{ma} \loopword{ọ} \loopword{bụ} \loopword{àgwà} \loopword{ndị} \loopword{yiri} \loopword{nke} \loopword{nna} \loopword{nna} \loopword{nwụrụ} \loopword{anwụ,} ma ihe ndị a nwere ike ịbụ n\textquotesingle{}ihi ọdịdị mkpụrụ ndụ ihe nketa ma ọ bụ mmetụta ọdịbendị. \loopword{Ụfọdụ} \loopword{ndị} \loopword{na-akọ} \loopword{akụkọ} \loopword{banyere} \loopword{ihe} \loopword{ndị} \loopword{dị} \loopword{ka} \loopword{ọdịdị} \loopword{anụ} \loopword{ahụ} \loopword{ma} \loopword{ọ} \loopword{bụ} \loopword{àgwà} \loopword{ndị} \loopword{yiri} \loopword{nke} \loopword{nna} \loopword{nna} \loopword{nwụrụ} \loopword{anwụ,} \dots} \\
\specialrule{0.4pt}{7pt}{7pt}
\textbf{LLaMA 3.1 8B}\hfill{\footnotesize\color{statgray}Type 3 \textperiodcentered{} question 1}\par\smallskip \loopbar{0.408}\hspace{6pt}{\footnotesize\textbf{41\%} of 147 words \textperiodcentered{} 3 repeats}\par\smallskip {\footnotesize\setlength{\fboxsep}{1pt}\dots{} ụmụ ọhụrụ: 1. \loopword{Ọnwụ} \loopword{ụmụ} \loopword{ọhụrụ} \loopword{bụ} \loopword{ihe} \loopword{a} \loopword{na-egosi} \loopword{mgbe} \loopword{ndị} \loopword{nne} \loopword{na} \loopword{nna} \loopword{gụrụ} \loopword{nwa} \loopword{aha} \loopword{ịma} \loopword{aka} \loopword{dị} \loopword{ka} \loopword{Ọnwụma.} Aha Ọnwụma bụ aha nke a na-egosi ọnwụ ụmụ ọhụrụ, na-eme ka ndị nne na nna na-achọ ịgụ aha ịma aka dị ka Ọnwụma mgbe ụmụ ha nwụrụ. 2. \loopword{Ọnwụ} \loopword{ụmụ} \loopword{ọhụrụ} \loopword{bụ} \loopword{ihe} \loopword{a} \loopword{na-egosi} \loopword{mgbe} \loopword{ndị} \loopword{nne} \loopword{na} \loopword{nna} \loopword{gụrụ} \loopword{nwa} \loopword{aha} \loopword{ịma} \loopword{aka} \loopword{dị} \loopword{ka} \loopword{Ọnwụma.} \dots} & \textbf{LLaMA 3.3 70B}\hfill{\footnotesize\color{statgray}Type 3 \textperiodcentered{} question 4}\par\smallskip \loopbar{0.291}\hspace{6pt}{\footnotesize\textbf{29\%} of 165 words \textperiodcentered{} 3 repeats}\par\smallskip {\footnotesize\setlength{\fboxsep}{1pt}\dots{} a na-egosi na \loopword{mmadụ} \loopword{niile} \loopword{nwere} \loopword{ebe} \loopword{ha} \loopword{dị} \loopword{iche,} \loopword{ebe} \loopword{ha} \loopword{na-akpata} \loopword{uru,} \loopword{na} \loopword{ebe} \loopword{ha} \loopword{na-aghọọ} \loopword{nwa.} Echiche a na-akpata ike site na echiche ọdịnala Igbo banyere ụmụ mmadụ na ebe ha dị na ụwa. Ọ na-akpọ mkpa ịhụ na \loopword{mmadụ} \loopword{niile} \loopword{nwere} \loopword{ebe} \loopword{ha} \loopword{dị} \loopword{iche,} \loopword{ebe} \loopword{ha} \loopword{na-akpata} \loopword{uru,} \loopword{na} \loopword{ebe} \loopword{ha} \loopword{na-aghọọ} \loopword{nwa.} \dots} \\
\specialrule{1.2pt}{7pt}{0pt}
\end{tabular}
\caption{Repetition loops in open-ended answers under Igbo prompting. Each excerpt is quoted verbatim from a model answer in the human-evaluation sample, with omitted text marked by an ellipsis; the repeated unit is highlighted, and the counts refer to the full stored answer. The annotator labelled all six answers as Model Breakdown and Repetition.}
\label{fig:repetition}
\end{figure*}

%% file: appendix_generated.tex
\section{Prompt Templates}
\label{app:prompt_templates}

This appendix reproduces the exact text of the twelve prompt templates. Placeholders in braces are filled from the dataset at run time. Each persona prompt consists of a persona introduction (Appendix~\ref{app:persona_intro}) followed by the zero-shot prompt of the same type and format (Appendix~\ref{app:zero_shot}); for Type~3 the introduction and the persona attributes are given in Igbo. A native-speaker linguist later identified minor phrasing issues in the Igbo instruction text; we retain the wording exactly as used during evaluation, for reproducibility.

\subsection{Zero-shot prompts}
\label{app:zero_shot}

\promptbox{Multiple choice, Type 1 (English instruction, English question)}{Answer in English only. Select exactly one option. Answer by typing only the letter (A, B, C, or D).\par\smallskip Question: \{question\}\par\smallskip Options:\\A. \{option\_a\}\\B. \{option\_b\}\\C. \{option\_c\}\\D. \{option\_d\}}
\promptbox{Multiple choice, Type 2 (English instruction, Igbo question)}{The question is in Igbo. Answer in Igbo only. Select exactly one option. Answer by typing only the letter (A, B, C, or D).\par\smallskip Question: \{question\}\par\smallskip Options:\\A. \{option\_a\}\\B. \{option\_b\}\\C. \{option\_c\}\\D. \{option\_d\}}
\promptbox{Multiple choice, Type 3 (Igbo instruction, Igbo question)}{Zaá n\textquotesingle{}asụsụ Igbo naanị. Họrọ naanị otu nhọrọ. Saá ya site nịde naanị akara (A, B, C, ma ọ bụ D).\par\smallskip Ajụjụ: \{question\}\par\smallskip Nhọrọ:\\A. \{option\_a\}\\B. \{option\_b\}\\C. \{option\_c\}\\D. \{option\_d\}}
\promptbox{Open-ended, Type 1 (English instruction, English question)}{Answer in English only. Provide a concise explanation in 2-4 sentences. Do not include any extra commentary.\par\smallskip Question: \{question\}}
\promptbox{Open-ended, Type 2 (English instruction, Igbo question)}{The question is in Igbo. Answer in Igbo only. Provide a concise explanation in 2-4 sentences. Do not include any extra commentary.\par\smallskip Question: \{question\}}
\promptbox{Open-ended, Type 3 (Igbo instruction, Igbo question)}{Zaá n\textquotesingle{}asụsụ Igbo naanị. Nye nkọwa dị mkpụmkpụ n\textquotesingle{}ime ahịrịokwu 2-4. Echefula itinye okwu ọ bụla ọzọ karịa azịza gị.\par\smallskip Ajụjụ: \{question\}}

\subsection{Persona introductions}
\label{app:persona_intro}

\promptbox{English introduction (Types 1 and 2)}{Imagine you are a \{marital\_status\} \{sex\} from \{state\} State, in South-Eastern Nigeria. You have completed \{education\} education and consider yourself part of \{social\_class\}. You have lived within Igbo cultural and community life throughout your life.\par\smallskip Others will read your response; your goal is to convince them that it reflects the genuine cultural knowledge and lived experience of the person described above.\par\smallskip Do not default to generalised Nigerian or African responses. Answer specifically from within Igbo cultural life as understood in \{state\} State.}
\promptbox{Igbo introduction (Type 3)}{Tụlee na ị bụ \{sex\} \{marital\_status\} si \{state\} Steeti, na Ndịda Ọwọwa Naịjirịa. Ị zụlichara \{education\} agụmakwụkwọ ma were onwe gị dị ka onye nke \{social\_class\}. Ị bi n\textquotesingle{}ime ọmenala na ndụ obodo Igbo n\textquotesingle{}oge niile n\textquotesingle{}ime ndụ gị.\par\smallskip Zaá ajụjụ a sitere n\textquotesingle{}echiche ahụ. Ndị ọzọ ga-agụ ihe i họọrọ; ebumnuche gị bụ ịkwenye ha na ọ bụ site n\textquotesingle{}ihe ọmụma ọmenala na ahụmahụ ndụ nke onye a kọwara n\textquotesingle{}elu.\par\smallskip Echefula ịza n\textquotesingle{}ụzọ izugbe nke Naịjirịa ma ọ bụ Afrịka. Zaá kpọmkwem site n\textquotesingle{}ime ndụ ọmenala Igbo dịka a maara ya na \{state\} Steeti.}

\subsection{Persona attributes}

The persona prompts combine each of the five personas below with each of the two states (Imo and Ebonyi), giving ten persona conditions. Each cell lists sex, marital status, education, and social class as inserted into the template.

\begin{table}[H]
\centering\footnotesize
\begin{tabular}{@{}L{0.18\columnwidth}L{0.35\columnwidth}L{0.35\columnwidth}@{}}
\toprule
\textbf{Persona} & \textbf{English (Types 1, 2)} & \textbf{Igbo (Type 3)} \\
\midrule
Young Teenage Boy & young boy; single; secondary school student; the youth & nwata nwoke; na-alụbeghị nwanyị; nwa akwụkwọ sekọndrị; ndị ntorobịa \\
\addlinespace
Married Woman & woman; married; primary school; the local trading class & nwaanyị; lụrụ di; ụlọ akwụkwọ praịmarị; ndị ahịa obodo \\
\addlinespace
Married Man & man; married; basic community; the rural farming class & nwoke; lụrụ nwaanyị; obodo; ndị ọrụ ugbo \\
\addlinespace
Widowed Woman & woman; widowed; vocational; the artisan class & nwaanyị; di ya nwụrụ; aka ọrụ; ndị aka ọrụ \\
\addlinespace
Professor & man; married; post-graduate (PhD); the academic middle class & nwoke; lụrụ nwaanyị; mahadum (PhD); ndị gụrụ akwụkwọ nke ọma \\
\bottomrule
\end{tabular}
\caption{Persona attributes inserted into the persona introductions.}
\label{tab:persona_attributes}
\end{table}